# Free Access to World News: Reconstructing Full-Text Articles from GDELT

**Andrea Fronzetti Colladon [1,*] and Roberto Vestrelli [2]**

1 Department of Civil, Computer Science and Aeronautical Technologies Engineering, Roma Tre University, Via della Vasca Navale 79, 00146 Rome, Italy
2 Department of Engineering, University of Perugia, Via G. Duranti 93, 06125 Perugia, Italy; roberto.vestrelli@dottorandi.unipg.it
* Correspondence: andrea.fronzetticolladon@uniroma3.it

**Abstract**

News data have become essential resources across various disciplines. Still, access to full-text news corpora remains challenging due to high costs and the limited availability of free alternatives. This paper presents a novel Python package (gdeltnews) that reconstructs full-text newspaper articles at near-zero cost by leveraging the Global Database of Events, Language, and Tone (GDELT) Web News NGrams 3.0 dataset. Our method merges overlapping n-grams extracted from global online news to rebuild complete articles. We validate the approach on a benchmark set of 2211 articles from major U.S. news outlets, achieving up to 95% text similarity against original articles based on Levenshtein and SequenceMatcher metrics. Our tool facilitates economic forecasting, computational social science, information science, and natural language processing applications by enabling free and large-scale access to full-text news data.

## 1. Introduction

Newspaper data present an extremely valuable resource, as demonstrated by their applicability across various research areas, spanning economics, finance, management, social sciences, computational science, and computer science. In economics and finance, news content has been used to predict stock price reversals [1] and to examine the impact of negative sentiment toward companies on their financial performance [2]. Other studies have used news to forecast financial markets [3–6] or to explain the cross-section of stock returns [4]. In the macroeconomic field, news has been employed to construct economic indicators. For example, Davis [7] developed an index to quantify global political uncertainty, demonstrating its correlation with recessions and changes in employment. Another wide branch of literature has examined news to investigate the impact of viral stories, motivated by the power of news to influence public opinion and amplify economic trends [8–11]. Communication scholars have long examined the media's ability to shape public perceptions [12,13] and influence which topics gain prominence in social and political discourse [14,15]. Social scientists have relied on the news to investigate protests [16–18], forecast elections [19], and evaluate the evolution of public opinion [20].

A growing body of research in information processing and retrieval emphasizes large-scale news analysis as a means to transform unstructured content into structured, actionable knowledge. In this context, news articles have become a foundational data source for tasks ranging from event detection to content recommendation. For example, Vossen et al. [21] developed a multilingual system that extracts episodic situational knowledge from news articles across four languages. Bouras & Tsogkas [22] explored enhanced clustering methods for organizing web news content using WordNet. Cleger-Tamayo et al. [23] contributed to personalized content delivery by proposing probabilistic models that recommend news based on user browsing profiles, demonstrating improved performance over standard baselines.

Today's advancements in language models have increased the demand for large-scale news datasets needed for training and validating these models. For example, 82% of the raw tokens used to train GPT-3 come from news data [24]. Nowadays, Large Language Models (LLMs) are employed for automated news analysis, ranging from sentiment analysis [25] to tasks such as automated news summarization [10].

Despite the extensive use of news data, a fundamental, under-addressed problem persists: most researchers lack affordable, flexible, and transparent access to full-text news articles at scale. Full-text access is often essential to preserve the information's semantic, contextual, and structural richness, which headlines, summaries, or metadata alone cannot adequately capture.

Recent studies have shown that n-gram representations remain effective for various natural language processing tasks, including text similarity detection and text classification. For example, Stefanovič et al. [26] showed that word-level n-grams can be successfully employed to assess document similarity when combined with classical similarity measures and self-organizing maps. More recently, Şen et al. [27] demonstrated that n-gram features can be leveraged to enrich word embeddings and improve text classification performance when integrated with graph convolutional networks. However, these approaches assume the availability of the original full-text documents as input, from which n-grams are subsequently extracted during the preprocessing stage. In both cases, n-grams serve as an intermediate or auxiliary representation derived from complete text, and model training, evaluation, and validation are all grounded in full-text corpora. As a result, these methods do not address the inverse problem of reconstructing coherent documents when only fragmented n-gram data are available and the original text is inaccessible. Moreover, most existing text similarity and classification pipelines are designed, trained, and benchmarked under the assumption of dense, contiguous textual input. Their performance and applicability therefore degrade substantially in settings where only sparse, disaggregated n-gram fragments are observed, such as in proprietary or privacy-constrained data releases.

Despite the clear demand, obtaining full-text corpora remains a challenging task. Some prior work has relied on web scraping, which may face legal restrictions and typically only provides access to the most recent articles. Other researchers have utilized data providers such as Factiva, LexisNexis, and Event Registry, which offer full-text access; however, using these platforms can be costly. Free alternatives exist, such as datasets available on Kaggle or GitHub, but they often suffer from limitations in terms of completeness and transparency regarding data collection. Another option consists of relying on providers that do not directly offer news text but provide pre-calculated text-based metrics, such as sentiment scores and other linguistic variables. Notable examples include RavenPack. However, the lack of full-text access has limited the ability to customize analyses and has prevented researchers from verifying the reliability of the provided metrics [28]. As a result, a clear gap exists between the growing demand for large-scale news text and the practical ability of researchers to obtain it.

Table 1 compares alternative approaches to accessing news content along four key dimensions. *Full-text Access* indicates whether the approach provides complete article text rather than headlines, summaries, or derived features. *Cost* reflects the typical financial burden for academic or institutional users, ranging from low (free or minimal infrastructure costs) to high (subscription or licensing fees). *Custom Text Analysis* captures the extent to which users can freely apply their own text-processing, linguistic, or machine-learning methods to the underlying content, as opposed to being restricted to precomputed metrics. Finally, *Legal Transparency* refers to the clarity and robustness of the legal framework governing data access and reuse, including licensing terms and compliance with copyright restrictions. The comparison highlights that, unlike existing solutions, our approach combines open access, full-text reconstruction, scalability, and reproducibility, thereby addressing a gap that prior work has not covered.

**Table 1.** Comparison of existing approaches for accessing news text and the proposed method.

| Approach | Full-Text Access | Cost | Custom Text Analysis | Legal Transparency |
|---|---|---|---|---|
| Proprietary news databases (e.g., Factiva, LexisNexis) | Yes | High | High | High |
| Event Registry | Yes | Medium | High | High |
| Web scraping-based datasets | Yes | Low | High | Low |
| Open news datasets (Kaggle, GitHub) | Partial | Low | Medium | Medium |
| Derived-text providers (e.g., RavenPack) | No | High | Low | High |
| Proposed GDELT-based reconstruction (this study) | Yes | Low | High | High |

From an algorithmic perspective, prior research has explored text reconstruction from n-grams under substantially different problem formulations and objectives. Gallé and Tealdi [29] studied the theoretical limits of document reconstruction from unordered n-gram multisets, with the explicit goal of identifying the longest substrings whose presence in the original text is guaranteed across all possible reconstructions. Their method intentionally avoids recovering a specific original document. Instead, it focuses on irreducible representations of ambiguity using de Bruijn graph reductions. In contrast, our work focuses on the faithful reconstruction of individual news articles by leveraging positional metadata associated with each n-gram.

A different approach was presented by Srinivas et al. [30], who addressed text file recovery in a digital forensics setting by using n-gram language models to probabilistically infer successor relationships between fragmented text blocks. Their objective was to recover a plausible fragment ordering rather than to reproduce the original document verbatim, and reconstruction quality was evaluated at the level of successor-edge accuracy rather than textual fidelity. Moreover, their approach operated without access to positional or document-level metadata. By contrast, our method explicitly leverages positional offsets and validates reconstruction quality using character- and sequence-level similarity metrics against a known ground truth, enabling systematic assessment of reconstruction accuracy rather than plausibility alone.

Accordingly, our study introduces a novel approach to reconstruct full-text news articles from fragmented n-gram data that are freely available through the GDELT Web News NGrams 3.0 dataset. Unlike prior work that either relies on proprietary sources or accepts fragmented text as a limitation, our approach demonstrates that it is possible to rebuild coherent and usable news articles by systematically assembling overlapping news fragments using positional information. The significance of this contribution lies not only in the technical reconstruction itself, but also in its practical implications: it enables researchers to create large-scale, customizable, and low-cost news corpora while avoiding legal restrictions and financial barriers.

To clearly delineate the novelty and value of this study, the main contributions of the article are summarized as follows:

- Methodological contribution: We propose a novel reconstruction methodology that assembles full-length news articles from unordered and overlapping n-gram fragments using a maximum-overlap strategy constrained by positional metadata.
- Algorithmic implementation: We design and release an open-source, Python-based package that operationalizes this methodology, enabling scalable reconstruction, filtering, and deduplication of large-scale news corpora from the GDELT Web News NGrams 3.0 dataset.
- Empirical validation: We validate the reconstructed articles against a benchmark full-text corpus obtained from EventRegistry, demonstrating high textual fidelity using sequence-sensitive metrics (Levenshtein and SequenceMatcher).
- Broader impact: By enabling near-zero-cost access to structured news text at scale, this work lowers barriers to entry for data-intensive research and supports more inclusive participation in computational social science and related fields.

*Research Objectives*

Previous studies have highlighted the critical role of news data in fields such as economics, finance, social sciences, and computational linguistics. Yet, researchers continue to face substantial barriers in accessing complete and reliable news corpora. Building on these challenges, our study seeks to develop a scalable, cost-effective approach for reconstructing full-text news articles from fragmented information in the GDELT Web News NGrams 3.0 dataset. By doing so, we aim to enhance the accessibility of large-scale textual data for empirical research, while overcoming the limitations posed by commercial or free, yet incomplete, datasets. Our research objective is to design, implement, and validate a Python-based package capable of rebuilding coherent news articles from n-gram fragments. Specifically, this study addresses two central research questions. First, to what extent can full-text news articles be accurately reconstructed from the fragmented n-grams provided by the GDELT Web News NGrams 3.0 dataset? Second, can the reconstructed corpus be a viable alternative to costly full-text news datasets?

## 2. GDELT Overview

GDELT is a vast, open dataset that captures and analyzes news media worldwide in real-time. Its system continuously monitors print, broadcast, and online news sources in over 100 languages, extracting structured information on global events. By leveraging advanced NLP techniques, GDELT translates, classifies, and organizes media content to create a comprehensive record of global events. These events are then encoded using the Conflict and Mediation Event Observations (CAMEO) framework, which categorizes interactions between actors, identifies event types, and assigns geographic and temporal markers to each event. Studies using GDELT span various domains, including the spread of misinformation and fake news during the COVID-19 pandemic [31,32]; global news coverage of disasters and refugee crises [33,34]; the influence of fake news on the online media ecosystem during the 2016 US Presidential election [35]; and the relationship between news framing and socio-political events [36]. Additionally, GDELT has been used to analyze protests, revolutions, and other forms of civil unrest [37–41], as well as state repression of such movements [42]. Furthermore, researchers have leveraged GDELT to examine institutional and civil society responses to the COVID-19 pandemic [43–45]. Studies using

GDELT data have been published in some of the most prestigious academic journals, including Science [46], Scientific Reports [47], The Quarterly Journal of Economics [48], and Organization Science [49].

*The GDELT News Ngrams Dataset*

The Python package (version 1.0.0) provided alongside this paper leverages one of the many datasets provided by GDELT: the Web News NGrams 3.0 Dataset. This dataset includes unigrams and a brief contextual snippet for each unigram, providing context for each term. The data is extracted from worldwide online news sources monitored by GDELT since 1 January 2020, and is updated every 15 min. According to GDELT, the dataset covers 42 billion words of news content in 152 different languages. Each entry in the dataset consists of a different "unigram" and has several metadata fields, including a brief contextual snippet that shows the unigram's surrounding context. These snippets are short textual fragments that precede and follow the unigram, enabling relevance filtering and context determination. Typically, they include up to seven words for space-segmented languages or an equivalent amount of semantic information for scriptio continua languages. In addition to these contextual words, GDELT provides other metadata, such as the URL of the original article, the date when the article was detected by GDELT, the language of the article, and the language type, which can take one of two values:

- 1: Indicates that the language uses spaces to segment words, meaning that n-grams correspond to words.
- 2: Indicates that the language follows a scriptio continua structure (e.g., Chinese or Japanese), where words are not separated by spaces, meaning that n-grams correspond to characters.

Each entry's unigram is assigned a position indicator, represented by a decile value based on the portion of the article where it appears. This allows researchers to assess whether a given word is mentioned at the beginning or end of an article. Despite its strengths, the News NGrams dataset presents certain challenges. Its main limitation is the inability to access the whole corpus of news articles. To address this issue, this paper proposes a methodology implemented in an open-source Python package (gdeltnews), freely available on GitHub (https://github.com/iandreafc/gdeltnews (accessed on 23 November 2025)) and on pypi.org, which enables the reconstruction of news articles' text from a collection of n-grams. A quick-start guide is provided in Appendix A.

The Python code presented in this paper only handles type 1 languages. Still, we plan to extend it to type 2 in the future. Extending the proposed reconstruction methodology to type 2 languages (scriptio continua languages such as Chinese and Japanese) introduces several non-trivial technical challenges. In these languages, n-grams correspond to individual characters rather than clearly delimited words, which substantially increases ambiguity during fragment matching and sequence reconstruction. First, character-level n-grams may belong to multiple possible word boundaries, making it difficult to determine how adjacent fragments should be merged into a coherent textual sequence. Second, the overlap-based matching strategy used for space-delimited languages relies on word-level continuity, which cannot be directly applied when fragments overlap at the character level without explicit segmentation. Third, positional metadata alone may be insufficient to disambiguate competing reconstruction paths, as short character sequences can appear frequently across different contexts within the same article.

A feasible extension to type 2 languages would therefore require adapting the reconstruction pipeline in several ways. While such extensions are technically feasible, they would increase computational complexity and necessitate careful evaluation to strike a balance between reconstruction accuracy and scalability. For these reasons, the current implementation focuses on type 1 languages, while support for type 2 languages is left for future work.

## 3. Reconstructing News Text

This section describes the complete pipeline that enables users to build a full-text corpus of news articles from GDELT. The pipeline has three main steps: (i) data download, (ii) article reconstruction, and (iii) optional filtering and URL-level deduplication. The reconstruction stage (Step 2) is the methodological core and is described in detail in Section 3.3, including assumptions, design choices, computational considerations, and limitations.

### 3.1. Pipeline Overview and Data Model

Input format. The reconstruction process starts from GDELT Web News NGrams 3.0 min files. Each record is an n-gram observation with metadata such as the article URL, a language code, a position indicator (pos) that approximates where the snippet appears in the article (in deciles), and a keyword-in-context (KWIC) view split into three fields: pre (tokens before the n-gram), ngram (the focal term), and post (tokens after the n-gram). Because the dataset contains only local contexts rather than complete sentences, and does not provide a single global ordering of all contexts, reconstruction becomes an assembly task: longer text must be built by merging many short, partially overlapping fragments.

Output format. The pipeline outputs a set of CSV files. Each row corresponds to a reconstructed article and includes at least the reconstructed text and the URL, as well as (when available or selected by the user) date metadata and source. After optional filtering and URL-level deduplication, users can consolidate these outputs into a single CSV.

Figure 1 illustrates the end-to-end pipeline in which fragmented GDELT Web News NGrams from online news are algorithmically reconstructed into coherent full-text articles by assembling many short, partially overlapping contextual snippets.

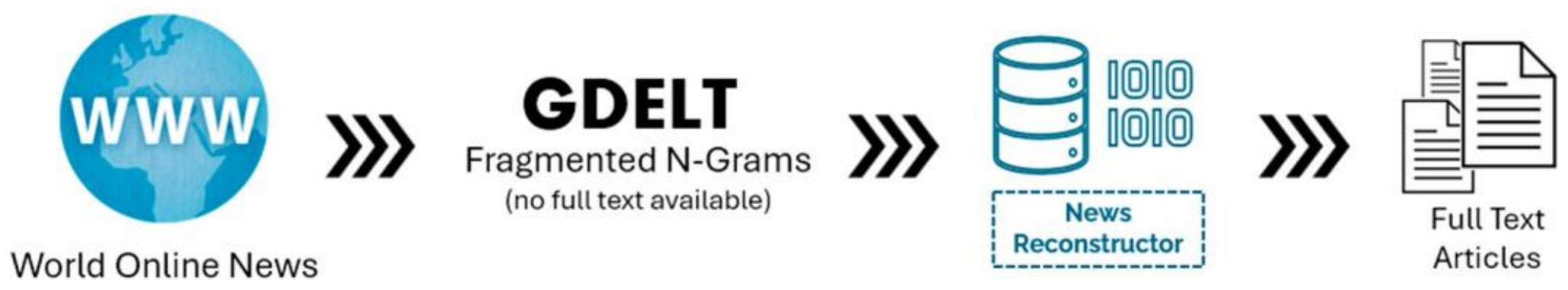

**Figure 1.** GDELT news reconstruction process.

### 3.2. Step 1: Downloading Minute-Level Web News NGrams Data

The pipeline takes as input a UTC start timestamp and an end timestamp. It then downloads all relevant files from GDELT's official repository for that time range. The package enumerates every minute in the (inclusive) range and constructs the corresponding filename using the format YYYYMMDDHHMMSS.webngrams.json.gz. Each file is retrieved over HTTP from the public GDELT Web News NGrams directory and saved to a local output folder. The result is a local directory of compressed JSON files, which becomes the input for Step 2.

### 3.3. Step 2: Reconstructing Articles from KWIC Fragments

In Step 2, the pipeline processes the downloaded files and reconstructs article text for each unique URL. Each input file is processed independently and produces one CSV output. This design supports parallel processing, incremental runs, and auditability, since users can trace each output file back to its source.

#### 3.3.1. Preprocessing: Grouping and Fragment Creation

For a given file, the software parses all records. It groups them by the URL field so that all records from the same article are processed together. For each record, it builds a text fragment by concatenating the three KWIC fields:

$$f = normalize(pre||ngram||post)$$

Normalization collapses repeated whitespace and trims leading and trailing spaces. Each fragment is therefore a short window of tokens centered on an observed n-gram occurrence.

As an illustrative case, take a single record with *pre* = "chip facilities to bypass environmental reviews that", *ngram* = "Commerce", and *post* = "Secretary Gina Raimondo has warned could take". The fragment is computed according to the formula above, which yields $f$ = "chip facilities to bypass environmental reviews that Commerce Secretary Gina Raimondo has warned could take".

#### 3.3.2. Artifact Removal Heuristic

We observed a recurring extraction artifact: sometimes the end of an article appears incorrectly attached to the beginning of the article inside the *pre* field, separated by the string "/". This is especially damaging for early fragments, because it can cause the overlap-based assembly to follow false transitions. To mitigate this issue, the implementation employs a conservative rule. If a fragment is likely to come from the beginning of the article (*pos* < 20) and it contains the delimiter "/", the fragment is split on "/", and the portion before the delimiter is discarded.

#### 3.3.3. Core Reconstruction Strategy: Maximum-Overlap Assembly with a Position Constraint

After fragments are created, reconstruction iteratively merges them into a single token sequence. The main signal for adjacency is token overlap at fragment boundaries. After normalization, fragments are tokenized based on whitespace. Given a current partial reconstruction $T$ and a candidate fragment $F$, we compute:

- The longest suffix of $T$ that matches the prefix of $F$ (append case), and
- The longest suffix of $F$ that matches the prefix of $T$ (prepend case).

The overlap length (the number of consecutive identical tokens) is used as the primary score to decide which fragment to merge next. When a merge occurs, overlapping tokens are included only once: the algorithm appends (or prepends) only the non-overlapping remainder.

Because *pos* provides only an approximate location estimate (in deciles), it cannot fully determine the correct order of fragments. However, it is still useful for preventing clearly implausible merges (for example, placing an end-of-article fragment before a beginning fragment). We therefore use *pos* as a constraint:

- Let $p_{\min}$ and $p_{\max}$ be, respectively, the minimum and maximum pos values already included in the current reconstruction.
- A new fragment may be appended only if its *pos* is not earlier than $p_{\max}$
- A new fragment may be prepended only if its *pos* is not later than $p_{\min}$

Among fragments that satisfy these constraints, the algorithm selects the fragment that yields the largest overlap with the current reconstruction boundary. If no remaining fragment both satisfies the position constraint and produces a positive overlap, reconstruction stops.

Reconstruction begins from a single seed fragment. In the current implementation, the seed is the fragment with the earliest *pos* within the URL group, which provides a natural left-to-right anchor. For reproducibility, ties (for example, equal overlap scores) are resolved deterministically, such as by stable ordering using *pos* and then fragment length or input order. As a result, the same input and parameters produce the same reconstructed output.

#### 3.3.4. Practical Filters and Parallelization

Step 2 provides two optional filters applied before reconstruction:

- Language filter: users can restrict processing to records with a specific language code, which reduces cross-language mixing
- URL filter: users can restrict processing to URLs containing one or more substrings (comma-separated), enabling source-specific corpora (for example, only articles from "nytimes.com").

For scalability, Step 2 supports multiprocessing. Since URL groups are independent after grouping, reconstruction can be parallelized across URLs (or batches of URLs) to use multiple CPU cores. The implementation generates one output CSV file per processed input file and deletes intermediate decompressed JSON files after processing to minimize disk usage.

### *3.4. Step 3: Final Filtering and URL-Level Deduplication*

Step 3 produces a final, analysis-ready dataset by consolidating per-minute CSV outputs and optionally applying semantic filtering and deduplication.

Boolean query filtering. Users can provide a Boolean query using AND, OR, NOT, parentheses, and quoted multi-word phrases. The software parses the query and evaluates it on each reconstructed article using case-insensitive substring matching. All matching rows across the CSV files are written to a temporary combined CSV with a standardized schema (Text, Date, URL, Source).

URL-level deduplication. To avoid overcounting and improve corpus completeness, the software groups rows by URL and retains a single representative row per URL. It retains the row with the longest reconstructed text, using length as a proxy for the completeness of the reconstruction. The final output is written to a single CSV file.

## 4. Data Collection and Validation

To validate our method, we require a benchmark corpus of original news articles to compare against the reconstructed articles generated by our pipeline. For this purpose, we use EventRegistry [50]. This widely adopted news aggregation platform offers access to full-text articles from various sources. We construct a benchmark dataset by downloading online articles published during the second half of December 2023 (15–31 December 2023) from several major U.S. news outlets: The New York Times (NYT), CNN, The Washington Post (TWP), The Wall Street Journal (WSJ), Bloomberg, and PRNewswire. We then collect all available articles from the GDELT Web NGrams 3.0 dataset for the same time window. By matching articles across the two datasets using their URLs, we construct a merged corpus of 2211 articles: 167 from Bloomberg, 10 from CNN, 389 from NYT, 493 from TWP, 252 from WSJ, and 900 from PRNewswire. The total number of GDELT entries, representing all n-grams, is 3,634,545.

EventRegistry provides full-text access at scale; however, it can introduce discrepancies relative to GDELT captures due to differences in retrieval time (article updates), variation in scraping and extraction (missing segments, paywall truncation, and boilerplate inclusion), and duplication. To reduce these sources of noise and make the validation more interpretable, we construct the comparison set by matching articles strictly by URL and applying a series of preprocessing and consistency controls.

First, to prevent duplicated content from influencing the results, we ensure that each URL appears only once in the evaluation corpus. When multiple records correspond to the same URL, we retain only a single instance. In addition to URL-level deduplication, we remove exact duplicate article bodies after normalization (identical token sequences), ensuring that syndicated or repeated content does not overweight the results. Second, we clean both reconstructed and EventRegistry texts using the same normalization pipeline (whitespace normalization, punctuation handling, and removal of non-substantive artifacts) prior to tokenization and similarity computation. Third, we compare EventRegistry and reconstructed texts under multiple evaluation regimes. We first report results on the full set of URL-matched pairs, which naturally includes cases where the two sources captured different versions of the same article or extracted different components (e.g., title, boilerplate, or truncated sections). We then repeat the evaluation on progressively stricter subsets defined by minimum token overlap. High-overlap pairs are more likely to reflect comparable extractions of the same underlying article text (and, when present, similar inclusion of titles or other fields). In contrast, lower-overlap pairs provide a realistic assessment of performance under cross-source noise and version mismatch.

To evaluate how well our method reconstructs the original text, we compare each GDELT-reconstructed article with the body of its corresponding EventRegistry article using two approaches: Levenshtein and Sequence Matching similarity. In this study, reconstruction accuracy is primarily defined in terms of textual fidelity, which refers to the extent to which the reconstructed article reproduces the original sequence of tokens and sentences from the source text. The Levenshtein and SequenceMatcher metrics were selected because they explicitly account for both token overlap and word order, which are central challenges when rebuilding articles from unordered and overlapping n-gram fragments. High values on these metrics indicate that large contiguous portions of the reconstructed text match the original article in both content and sequence.

At the same time, it is important to distinguish textual fidelity from full semantic equivalence. While sequence-based similarity metrics are sensitive to omissions, insertions, and reordering of text, they do not directly measure whether the interpretation or narrative meaning of an article is preserved. For example, the removal of a short but contextually important paragraph or a minor reordering of sentences may have a limited impact on similarity scores while affecting certain qualitative analyses. Conversely, small stylistic differences or boilerplate content may reduce similarity scores without materially changing meaning.

As a result, the reported similarity values should be interpreted as evidence that the reconstructed articles closely approximate the original texts in structure and lexical content, rather than as a guarantee of perfect semantic equivalence in all cases. This distinction is particularly relevant for applications that depend on fine-grained narrative structure, causal sequencing, or rhetorical framing.

First, we use the Ratio function of the Levenshtein package (https://pypi.org/project/python-Levenshtein/ (accessed on 23 November 2025)) to calculate a Levenshtein similarity metric given by the following formula:

$$Levenshtein\ Similarity = 1 - \frac{Indel\ Distance}{len(t_1) + len(t_2)}, \tag{1}$$

where $len(t_x)$ denotes the length of the text $x$. The Indel distance measures the minimum number of character insertions and deletions needed to transform one string into another. The formula gives a value of 0 when the two strings are completely different and a value of 1 if they are identical.

Second, we use the SequenceMatcher class from Python's difflib module (https://docs.python.org/3/library/difflib.html (accessed on 23 November 2025)) at the word level, to identify the longest matching sequences between two texts, giving more weight to consecutive matches and preserving token order. Specifically, SequenceMatcher finds the longest contiguous matching subsequence between two strings, then recursively repeats the process for the substrings before and after that match. This continues until no more matching blocks are found. The similarity ratio is then calculated as:

$$SequenceMatcher\ Similarity = (2 \times M)/TC \tag{2}$$

where $M$ is the total number of matching characters, and $TC$ is the total number of characters in both strings combined.

We selected these approaches over other similarity metrics because they account not only for the presence of shared words between two articles but also for the order in which those words appear. This is crucial for our goal: rebuilding the article body in the correct sequence from fragmented snippets. Since one of the core challenges of working with these text fragments is that they are unordered and partially overlapping, our method focuses specifically on reconstructing the correct order of words and sentences. For this reason, metrics that are sensitive to word order are especially suitable for validating the method.

As already mentioned, some differences between the EventRegistry and GDELT versions of the same article may be unrelated to our reconstruction algorithm. These can include differences in scraping methods (e.g., one may omit the title), updates or edits made to the article over time, or inconsistencies in how the content is segmented. To assess our method under different conditions, we conduct two analyses. First, we compute similarity metrics for all matched article pairs without any filtering. Second, we restrict the comparison to article pairs that share a minimum number of tokens (measured by Jaccard Similarity), regardless of order. The idea is that if two articles share most of their content, they likely represent the same version of the text and can serve as a stronger test of our method's ability to reconstruct token order.

Results are shown in Table 2. Without filtering, the average similarity between reconstructed and original articles is 0.75 (Levenshtein) and 0.73 (SequenceMatcher). When we filter for article pairs that share at least 60% of tokens, similarity rises to 0.92 for both metrics. At 70% token overlap, it increases further to 0.94 and 0.93. When using an 80% threshold of minimum token overlap, similarity reaches 0.96 and 0.95. These findings support the validity of our reconstruction method. Even in the presence of minor noise, or missing metadata, our approach reliably rebuilds the article structure with high fidelity. In cases where the original and reconstructed articles are very likely to reflect the same underlying version, the similarity is nearly perfect.

**Table 2.** Validation results.

| | | Common Tokens (%) | | |
|---|---|---|---|---|
| **Metric** | **No Filter** | **>60%** | **>70%** | **>80%** |
| Levenshtein Similarity | 0.75 | 0.92 | 0.94 | 0.96 |
| SequenceMatcher | 0.73 | 0.92 | 0.93 | 0.95 |

Acknowledging that decile assignments may vary across news sources, Table 3 reports median similarity metrics separately by outlet to assess whether reconstruction performance is source dependent. Using the same minimum token-overlap threshold as in Table 1 (60%), we consistently find high similarity across publishers, with values ranging from 0.91 to 1.00 (Bloomberg). These results suggest that the method's performance is robust across heterogeneous sources and that reconstruction quality remains stable despite potential cross-publisher differences in extraction and positional decile assignment.

**Table 3.** Validation by source.

| | Similarity Metric | |
|---|---|---|
| **Data Source** | **Levenshtein Similarity** | **SequenceMatcher** |
| Bloomberg | 1.00 | 1.00 |
| CNN | 0.97 | 0.97 |
| NYT | 0.97 | 0.97 |
| PRNewswire | 0.94 | 0.94 |
| TWP | 0.92 | 0.91 |
| WSJ | 0.92 | 0.93 |

Using the same reconstruction and matching procedure, we select Italian-language articles from "La Repubblica" and match them with their corresponding versions in EventRegistry. This process yields a final dataset of 79 matched articles. As in Table 1, we report similarity metrics both without filtering and under increasingly strict minimum token-overlap thresholds in Table 4. The results indicate strong reconstruction performance also in a non-English context. Without filtering, average similarity reaches 0.83 for both Levenshtein and SequenceMatcher metrics. When restricting the comparison to article pairs sharing at least 60% of tokens, similarity rises to 0.90, increasing further to 0.92 at the 70% threshold and 0.94 when requiring at least 80% token overlap. These findings suggest that the proposed reconstruction method can generalize well across languages and remain robust to linguistic differences, confirming its applicability beyond the English-language sources analyzed above. Future studies could extend our analysis by considering more languages.

**Table 4.** Validation analysis in a multilingual setting.

| | | Common Tokens (%) | | |
|---|---|---|---|---|
| **Metric** | **No Filter** | **>60%** | **>70%** | **>80%** |
| Levenshtein Similarity | 0.83 | 0.90 | 0.92 | 0.94 |
| SequenceMatcher | 0.83 | 0.90 | 0.92 | 0.94 |

We emphasize that high similarity scores should be interpreted as indicating near-verbatim textual reconstruction rather than perfect equivalence across all analytical perspectives. While the reconstructed articles are well-suited for applications that rely on aggregate textual features, statistical language patterns, or large-scale content analysis, applications requiring fine-grained discourse analysis or exact narrative framing may require additional validation or targeted filtering.

*Computational Feasibility*

To complement the reconstruction accuracy results and to demonstrate the practical feasibility of the proposed scheme, we report wall-clock runtimes for the main stages of the reconstruction pipeline using a real data extraction scenario. All experiments were conducted on standard commodity hardware, utilising a single Intel i9-13980HX CPU core, which provides a conservative baseline for computational performance.

We consider a three-hour time window from 5 May 2023 10:00:00 to 5 May 2023 13:59:00, which corresponds to 39 Web News NGrams 3.0 files from GDELT. Downloading the complete set of compressed files required 4 min and 22 s, corresponding to an average of approximately 6.77 s per file. This stage is I/O-bound and scales linearly with the length of the requested time window.

After downloading, we restricted processing to Italian-language articles (language code “it”). These files were reconstructed using the proposed overlap-based assembly algorithm with positional constraints, executed on a single CPU core. The total reconstruction time for these 39 files was 1 h, 8 min, and 9 s, corresponding to roughly 1.7 min per minute-level file. This experiment reflects a conservative baseline, as no parallelization was employed.

The final stage of the pipeline involves URL-level deduplication and consolidation, where a single representative reconstructed article is retained per URL. No keyword-based filtering was applied. This step required less than one second and produced a final corpus of 6842 unique reconstructed articles, indicating that the post-processing overhead is negligible relative to the reconstruction process. Table 5 summarizes the runtime and output statistics for each stage of the pipeline.

**Table 5.** Runtime and output statistics for the reconstruction pipeline.

| **Pipeline Stage** | **Input** | **Runtime** | **Output** |
|---|---|---|---|
| Data download | 39 files | 4 min 22 s | 39 compressed files |
| Article reconstruction | 39 files | 1 h 08 min 09 s | Reconstructed articles |
| Deduplication and consolidation | Reconstructed articles | <1 s | 6842 unique articles |

These results demonstrate that the proposed reconstruction scheme is computationally feasible on standard hardware. Runtime scales predictably with the number of files processed, and the dominant cost lies in the reconstruction stage. Importantly, reconstruction is performed independently at the URL level, allowing the process to be naturally parallelized across CPU cores. As a result, substantially lower wall-clock times can be achieved in practice using multiprocessing, making the approach suitable for large-scale corpus construction and empirical research workflows.

## 5. Discussion and Implications

The results of this study demonstrate that it is possible to reconstruct full-text newspaper articles from the fragmented information contained in the GDELT Web News NGrams 3.0 dataset with high accuracy. By comparing reconstructed articles to original texts from EventRegistry, we achieve average similarity scores of up to 95% for article pairs with high token overlap. These findings confirm the feasibility of using large-scale n-gram datasets to approximate original news content at near-zero cost, thus providing an alternative to traditional subscription-based full-text news services.

The articles reconstructed by our method offer a practical, scalable alternative to proprietary full-text news datasets for a wide range of empirical applications. In particular, tasks such as sentiment analysis, topic modeling, event detection, economic forecasting, and language model training primarily depend on the presence and ordering of lexical content at scale. For these use cases, the high sequence-based similarity scores observed in our validation suggest that reconstructed articles can function as effective substitutes for original full-text sources.

At the same time, certain research applications place stronger demands on exact narrative sequencing, rhetorical emphasis, or the presence of specific contextual passages (e.g., qualitative discourse analysis or close reading of framing effects). In such cases, even minor omissions or reordering of fragments may have disproportionate analytical consequences. Researchers employing reconstructed articles for these purposes should therefore consider additional validation steps, such as semantic similarity checks or manual inspection of subsets of the data.

Compared with previous research that has relied on either full-text proprietary databases or pre-calculated sentiment and other text-mining scores without direct access to the text, our method offers a valid alternative. Researchers can gain access to structured article text without incurring high licensing fees while still retaining control over the text analysis process. This work demonstrates that even highly granular data, such as unigrams and their local contexts, can be effectively recombined to obtain full article bodies. A key finding is that the reconstruction method maintains good fidelity to the original article structure even in the presence of noise, missing fragments, and variations in scraping quality. This robustness suggests that the approach can be useful not only for research applications that require general semantic content (such as topic modeling and sentiment analysis) but also for more sensitive tasks where word order and coherence matter, such as narrative analysis or event extraction.

To the best of our knowledge, this is the first study to demonstrate that article-level representations can be effectively reconstructed from low-level textual fragments. This finding broadens the possibilities for research using partial text datasets, suggesting that with careful algorithmic design, researchers are not necessarily constrained to proprietary full-text corpora when conducting sophisticated natural language processing analyses. Our method also complements work in computational social science that increasingly relies on alternative big data sources [51–53]. Practically, the availability of a nearly-zero-cost tool for reconstructing large-scale news text datasets has significant implications. Researchers and practitioners, particularly those without access to expensive subscriptions, can now conduct empirical studies across economics, management, political science, and other fields of research with relatively few resource constraints. This democratizes access to data-intensive research opportunities and fosters more inclusive academic participation. Moreover, the tool we presented can be adapted to specific use cases by filtering news by language, source URL, or temporal range, allowing for tailored datasets to support targeted research projects. It also offers a foundation for the development of real-time news monitoring systems, given GDELT's frequent data updates.

## 6. Conclusions

In this paper, we introduced a method for reconstructing the full body of news articles using the GDELT's Web News NGrams 3.0 dataset. Our open-source Python package (gdeltnews) allows researchers to generate structured, large-scale newspaper text data at nearly zero cost, facilitating a wide range of applications in economics, social sciences, and natural language processing. The codebase, available at https://github.com/iandreafc/gdeltnews (accessed on 23 November 2025)), is maintained and designed to be modular and extensible. For example, it can be easily adapted to include filters based on specific news sources by leveraging the URL field provided by GDELT. This makes the tool flexible for targeted research use cases. However, the approach is not without limitations. Notably, article titles are not always included in the GDELT dataset. This limits the metadata available for organizing and referencing the reconstructed articles. Additionally, the single-process version of our code can be slow, especially when processing large volumes of data. While this is acceptable for exploratory or small-scale research, future improvements could involve automating the pipeline to run at regular intervals—e.g., every 15 min in sync with GDELT's update frequency—to enable near real-time processing. Lastly, we provide a code version that runs in parallel to address the speed problem. Further improving the computational efficiency of our code to handle near-real-time processing at scale would open new avenues for dynamic event detection and real-time analytics. Looking ahead, we plan to extend the code to support languages that do not use space-based word segmentation (e.g., Chinese or Japanese) and to improve the code's efficiency and precision. Future work could also extend our method and integrate mechanisms for identifying and retrieving article titles through external matching algorithms. These extensions aim to support the open dissemination of information and facilitate free access to high-quality research materials for scholars worldwide.

**Author Contributions:** A.F.C.: Conceptualization; Data curation; Formal analysis; Methodology; Software; Supervision; Validation; Visualization; Writing—original draft; Writing—review and editing. R.V.: Conceptualization; Data curation; Formal analysis; Methodology; Software; Validation; Visualization; Writing—original draft; Writing—review and editing. All authors have read and agreed to the published version of the manuscript.

**Funding:** This research received no external funding.

**Institutional Review Board Statement:** Not applicable.

**Informed Consent Statement:** Not applicable.

**Data Availability Statement:** The code used is publicly available on GitHub. The news data sourced from GDELT can be downloaded from its official website. The news data obtained from Event Registry cannot be shared due to copyright and licensing restrictions.

**Conflicts of Interest:** The authors declare no conflicts of interest.

## Appendix A.

*Appendix A.1. Quick-Start Guide*

This appendix provides a quick-start guide for using the gdeltnews Python package. The package can be installed from PyPI:

```
pip install gdeltnews
```

*Step 1*. The first step consists of downloading GDELT Web NGrams ZIP files for a specified time window. The start and end timestamps must be provided in ISO 8601 format. The destination directory must be specified via the "outdir" parameter. If "decompress" is set to True, downloaded files are automatically decompressed; otherwise, they are stored in compressed form (using ZIP files is recommended, since Step 2 processes them directly without requiring prior decompression).

```
from gdeltnews.download import download
download(
    "2025-11-25T10:00:00",
    "2025-11-25T13:59:00",
    outdir="gdeltdata",
    decompress=False,
)
```

*Step 2*. Full news articles are reconstructed from Web NGrams. To ensure correct multiprocessing behavior, this step should be executed as a standalone Python script rather than within a Jupyter notebook. Articles can be filtered by language and by source using a list of URL substrings. Setting these parameters to None turns off filtering. The processes argument controls the number of parallel worker processes.

```
from multiprocessing import freeze_support
from gdeltnews.reconstruct import reconstruct
def main():
    reconstruct(
        input_dir="gdeltdata",
        output_dir="gdeltpreprocessed",
        language="it",
        url_filters=["repubblica.it", "corriere.it"],
```

```
        processes=10,
    )
if __name__== "__main__":
    freeze_support()
    main()
```

*Step 3*. Reconstructed articles are filtered using a Boolean query, deduplicated, and merged into a single CSV file. The query supports the logical operators AND, OR, and NOT.

```
from gdeltnews.filtermerge import filtermerge
filtermerge(
    input_dir="gdeltpreprocessed",
    output_file="final_filtered_dedup.csv",
    query="((elections OR primaries) AND (democrats OR "'party')) AND NOT republicans"
)
```

*Appendix A.2. Graphical Version*

The software also includes a graphical user interface. Users interested in this option can run the main.py script located in the GUI folder of the GitHub repository after installing all required dependencies. The GUI reproduces the three steps described above in separate tabs labeled "Download", "Reconstruct", and "Filter/Merge". Screenshots of the interface are shown in Figure A1.

Both the command line and graphical versions of the software may be updated over time. For the most recent version, documentation, and usage instructions, users should refer to the official GitHub repository (https://github.com/iandreafc/gdeltnews (accessed on 23 November 2025)).

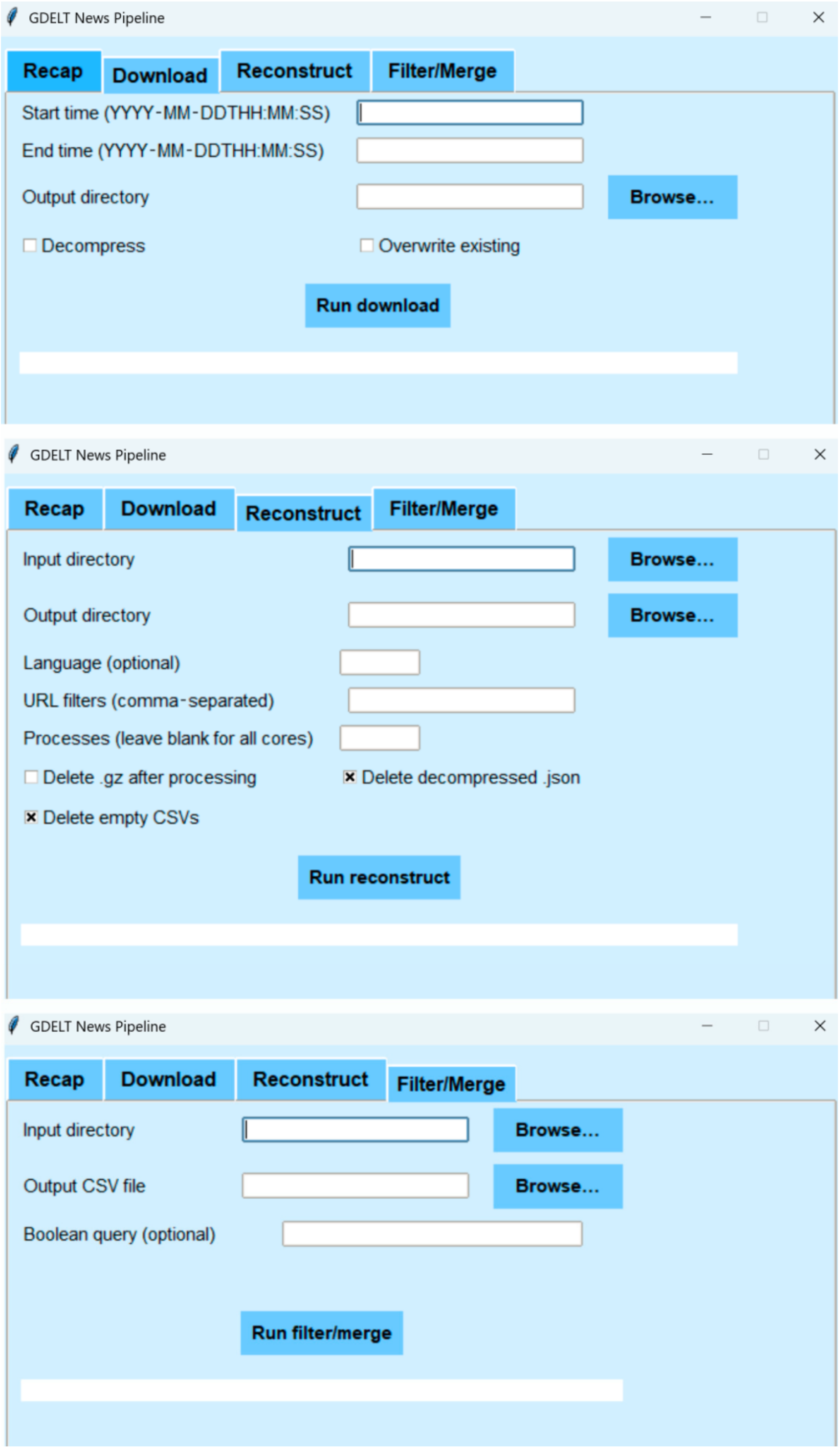

**Figure A1.** Graphical version.